\documentclass[10pt,journal,compsoc]{IEEEtran}

\IEEEoverridecommandlockouts
\usepackage{amsmath,amssymb,amsfonts}
\usepackage{algorithm,float}
\usepackage{algorithmic}
\usepackage{graphicx}
\usepackage{textcomp}
\usepackage{xcolor}
\usepackage{subcaption}
\usepackage{stfloats}
\usepackage{bm,bbm}
\newtheorem{theorem}{Theorem}

\newtheorem{assumption}{Assumption}
\newtheorem{remark}{Remark}

\usepackage[bottom]{footmisc}
\usepackage{tablefootnote} 
\usepackage{url}
\usepackage{balance}
\usepackage{lastpage}
\usepackage{enumitem,kantlipsum}
\usepackage{color,soul}  
\graphicspath{{../figures/}}
\newcommand{\norm}[1]{\left\lVert#1\right\rVert} 
\def\BibTeX{{\rm B\kern-.05em{\sc i\kern-.025em b}\kern-.08em
    T\kern-.1667em\lower.7ex\hbox{E}\kern-.125emX}}
 
\DeclareMathOperator*{\argmin}{arg\,min}

\usepackage{scalerel,stackengine}
\stackMath
\newcommand\reallywidehat[1]{%
\savestack{\tmpbox}{\stretchto{%
  \scaleto{%
    \scalerel*[\widthof{\ensuremath{#1}}]{\kern-.6pt\bigwedge\kern-.6pt}%
    {\rule[-\textheight/2]{1ex}{\textheight}}
  }{\textheight}%
}{0.5ex}}%
\stackon[1pt]{#1}{\tmpbox}%
}
\ifCLASSOPTIONcompsoc
  \usepackage[nocompress]{cite}
\else
  \usepackage{cite}
\fi

\ifCLASSINFOpdf
\else
\fi

\begin{document}

\title{Asynchronous Parallel Incremental Block-Coordinate Descent for Decentralized Machine Learning}

\author{
    	Hao~Chen,~\IEEEmembership{Student~Member,~IEEE,}
    	Yu~Ye,
	    Ming~Xiao,~\IEEEmembership{Senior~Member,~IEEE }
	    and~Mikael~Skoglund,~\IEEEmembership{Fellow,~IEEE}
\IEEEcompsocitemizethanks{\IEEEcompsocthanksitem H. Chen, Y. Ye, M. Xiao and M. Skoglund are with the School of Electrical Engineering and Computer Science, Royal Institute of Technology (KTH), 100 44 Stockholm, Sweden.
 	\protect\\
E-mail: \{haoch, yu9, mingx, skoglund\}@kth.se
}
\thanks{Manuscript received xx xx, xx; revised xx xx, xx.}}

\markboth{Journal of \LaTeX\ Class Files,~Vol.~xx, No.~xx, xx~xx}%
{Shell \MakeLowercase{\textit{et al.}}: Bare Demo of IEEEtran.cls for Computer Society Journals}

\IEEEtitleabstractindextext{%
\begin{abstract}
Machine learning (ML) is a key technique for big-data-driven modelling and analysis of massive Internet of Things (IoT) based intelligent and ubiquitous computing. For fast-increasing applications and data amounts, distributed learning is a promising emerging paradigm since it is often impractical or inefficient to share/aggregate data to a centralized location from distinct ones. 
This paper studies the problem of training an ML model over decentralized systems, where data are distributed over many user devices and the learning algorithm run on-device, with the aim of relaxing the burden at a central entity/server. 
Although gossip-based approaches have been used for this purpose in different use cases, they suffer from high communication costs, especially when the number of devices is large.
To mitigate this, incremental-based methods are proposed.
We first introduce incremental block-coordinate descent (I-BCD) for the decentralized ML, which can reduce communication costs at the expense of running time.
To accelerate the convergence speed, an asynchronous parallel incremental BCD (API-BCD) method is proposed, where multiple devices/agents are active in an asynchronous fashion.
We derive convergence properties for the proposed methods.
Simulation results also show that our API-BCD method outperforms state of the art in terms of running time and communication costs.
\end{abstract}

}

\maketitle
\IEEEdisplaynontitleabstractindextext
\IEEEpeerreviewmaketitle

\begin{figure} [t] 
	\begin{center}
		\centerline{\includegraphics[width=78 mm]{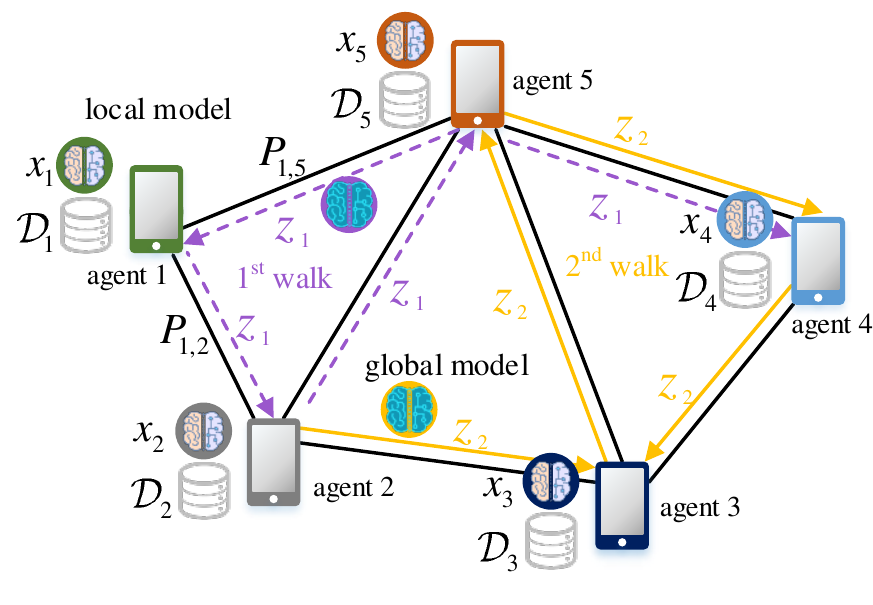}}
		\caption{An example of API-BCD approach on decentralized systems.}
		\label{fig:system}
	\end{center}
	\vskip -0.4in
\end{figure}

\IEEEraisesectionheading{\section{Introduction}
\label{sec:intro}}
{\color{black}
\IEEEPARstart{R}{ecently}, Internet of Things (IoT) facilitated a lot of compelling applications in a wide range of fields, such as augmented and virtual reality (AR/VR) \cite{ar_vr}, smart grids \cite{smart_grid}, drones \cite{drones}, etc.
Meanwhile, the fast development of machine learning (ML) has spurred many new applications and functions.
For instance, autonomous driving has drawn numerous attention from both academia and industry \cite{automomous_driving}.
Although ML techniques significantly improve the performance of IoT-based applications, massive data generated from these applications are often collected and stored in a distributed manner.
This induces excessive costs with regard to communication, computational and memory resources if the data are gathered, stored and analyzed in a central server/node.
To address these challenges, distributed ML is proposed as one promising emerging paradigm with the advent of powerful personal devices and edge computing since allocating the learning process close to the data sources or several workstations is a natural way of scaling up learning algorithms, especially analyzing abundant data for large-scale learning in massive IoT environments \cite{peteiro2013survey}.

Generally, decentralized ML refers to the scenario of learning a common pattern of interest among multiple devices/agents without a central parameter server (PS).
Here, ML is defined over an undirected and connected network $\mathcal{G}=(\mathcal{N},\mathcal{E})$, where $\mathcal{N}=\{1,...,N\}$ is the set of agents and $\mathcal{E}$ represents the set of bidirectional communication links, as depicted in Fig. \ref{fig:system}. 
Essentially, it can be formulated in the following form:
\begin{equation}\label{eq1}
\min_{x}~ \sum_{i=1}^{N}f_i(x),
\end{equation}
where $f_i: \mathbbm{R}^p\to\mathbbm{R}$ is the locally private loss function of agent $i$, and all agents aim to learn a common model parameterized by $x\in\mathbbm{R}^p$. Our goal is to design frameworks of incremental based decentralized learning algorithms, which enable agents to collaboratively learn a shared global model through on-device computation and information sharing among neighbours without relying on a PS. 
}

\textbf{Related works:}
With the rapid popularity of decentralized ML, for instance, federated learning coined by Google in 2016 \cite{fedavg} , it has become an attractive research direction in recent years \cite{li2020federated, hsieh2020non, xin2020decentralized, 8750869, chen2021federated}.
In the literature, for solving problem (\ref{eq1}), many decentralized algorithms have been developed.
Most of them are gossip-based and the well-known algorithms include decentralized gradient descent (DGD) \cite{DGD}, EXTRA \cite{EXTRA}, distributed alternating direction method of multipliers (D-ADMM) \cite{DADMM}, where each agent is allowed to exchange model parameters with a subset of or even all their direct neighbours at each iteration.
Although these approaches have good convergence rates, communication costs become extremely high, especially for large-scale ML tasks such as systems with slow or unstable connections in distributed ML \cite{2017federated}. 

In another line of work, an incremental-based learning framework has been recognized as an alternative technique to alleviate the communication burden, which activates one agent and on link at each iteration whilst keeping all other agents and links idle.
The existing schemes, for instance, random-walk ADMM (WADMM) \cite{WADMM_wotao}, walk proximal gradient (WPG) \cite{WPG}, parallel randomm walk ADMM (PW-ADMM) \cite{PWADMM}, are typical incremental methods.
Specifically, an agent is selected according to a Markov chain and the order pattern of activated agents is called random walk in WADMM \cite{WADMM_wotao}, while multiple random walks of agents are allowed simultaneously in PW-ADMM \cite{PWADMM}. 
This choice of clients can be further deterministic and the updating order of agents in WPG follows a predetermined cycle such as a Hamiltonian cycle \cite{WPG}.
However, in these methods, accuracy is sacrificed to achieve lower communication costs.

{\color{black}
\textbf{Our contributions:} Motivated by the above observations,
we will study decentralized ML established on the incremental framework. 
By generalizing the method in \cite{WPG}, we investigate the possibility of asynchronous parallel incremental block-coordinate descent (BCD) for achieving both reduced communication consumption and accelerated algorithm convergence compared with state-of-the-art approaches.
In this article, the main contributions to this study are as follows.
\begin{itemize}
\item We train a global model in a decentralized system by the incremental BCD (I-BCD) method that is first presented for communication efficiency, at the expense of running time.

\item To speed up algorithm convergence, an asynchronous parallel incremental BCD (API-BCD) method is proposed. In addition, a variant of API-BCD is discussed for a reduced computational complexity.

\item We provide both theoretical and experimental analyses for the provided approaches. 
Simulation results reveal that the proposed API-BCD method brings improved performance in terms of running time and communication costs compared with state-of-the-art methods. 
\end{itemize}

The remainder of this paper is organized as follows. 
Section \ref{sect:2} introduces the incremental BCD framework. 
In Section \ref{sect:3}, the proposed asynchronous parallel incremental learning scheme is derived.
The convergence properties of the provided approaches are discussed in Section \ref{sec:convergence}. 
Afterwards, we investigate the performance of the proposed algorithms in Section \ref{sec:4}.
Finally, the paper is concluded in Section \ref{sect:5}. 
} 
 
{\color{black}{ 
\section{Incremental Block-Coordinate Descent Framework} \label{sect:2}
We begin with the basic framework of incremental BCD (I-BCD) method for decentralized ML.
As illustrated in Fig. \ref{fig:system}, we consider a decentralized learning system consisting of multiple agents, where a common pattern of interest is collaboratively trained without a central PS.
Denoting by $x_i \in \mathbbm{R}^p$ the local learning model in agent $i\in\mathcal{N}=\{1,...,N\}$, $\bm{x} =  \{ x_1,...,x_N \} $ and $z \in \mathbbm{R}^p$ the global model, problem (\ref{eq1}) is equivalently expressed as the following constrained optimization problem
\begin{equation}\label{p1} 
    \min_{\bm x,z}~ \sum_{i=1}^{N}f_i(x_i); \quad s.t.~ z=x_i,~i\in\mathcal{N}, 
\end{equation}
where each local loss $f_i(x_i)$ can be calculated by $f_i(x_i) = \frac{1}{d_i} \sum_{l=1}^{d_i} \ell_i(x_i; \xi_{i, l})$ and $\ell_i(\cdot)$ is a loss function given its local training data $\mathcal{D}_i=\{{{\xi}_{i,l}}\}_{l=1}^{d_i}$.
Following \cite{6425938, 7178513, bajovic2017newtonlike}, {\color{black}{problem (\ref{p1}) is further reformulated as follows:}}
\begin{equation}\label{p2}
    \min_{\bm x,z}~ \mathcal{F}(\bm x,z) := \sum_{i=1}^{N}f_i(x_i) + \frac{\tau}{2}\sum_{i=1}^{N}\|x_i-z \|^2,
\end{equation}
in which $\mathcal{F}(\bm x, z) $ is an auxiliary function.
Here, constraints of problem (\ref{p1}) are combined into objective (\ref{p2}) as penalty terms with parameter $\tau$. 
Clearly, the penalty parameter $\tau$ influences the relation between problems (\ref{p1}) and (\ref{p2}), where a larger $\tau$ implies better agreement between the problems.
As stated in \cite{WPG}, if a central PS exits, we get the $k$-th iteration in client $i$ as follows
    \begin{align} 
        &x_i^{k+1} := \argmin_{x_i}f_i(x_i) + \frac{\tau}{2}\|x_i - z^k \|^2,~i \in \mathcal{N};\label{eq:eq4}\\
        &z^{k+1}:=\frac{1}{N}\sum_{i=1}^{N} x_i^{k+1}. \label{eq:eq5}
    \end{align}
It is observed that the PS collects local models from all agents via (\ref{eq:eq5}).
{\color{black}To solve (\ref{p2}) in a collaborative manner, incremental-based approach is one alternative, where global model $z$ can be updated sequentially in a succession of agents over the network.
Essentially, at each iteration $k$, agent $i$ that receives $z^k$ updates it with current local data $\mathcal{D}_i$, followed by sending $z^{k+1}$ to a neighboring agent.}
Following \cite{WADMM_wotao}, and under the initialization
\begin{align}
    z^{0}:=\frac{1}{N}\sum_{i=1}^{N} x_i^{0},
\end{align}
the I-BCD approach is
given as
 \begin{align}
        &x_i^{k+1} := \left\{\begin{aligned}&\argmin_{x_i}f_i(x_i) + \frac{\tau}{2}\|x_i - z^k \|^2,~i=i_k;\\
        &x_i^{k },~\text{otherwise};\label{eq:is_x}
        \end{aligned}\right. \\
       &z^{k+1}:=  z^k + \frac{1}{N}( x_{i_k}^{k+1} - x_{i_k}^k) .  \label{eq:is_z}
    \end{align}
Note that only $x_{i_k}$ of active agent $i_k$ gets updated, whilst all the other local models in $\{x_{j} |j \neq i_k, j \in \mathcal{N}\}$ stay unchanged.
In this method, the selection rule of agent $i_k$ over the iterations can be in a predetermined circulant pattern (e.g. Hamiltonian cycle in \cite{WPG, privacy_iadmm, codeadmm}) or a randomized pattern (e.g., Markov chain \cite{WADMM_wotao, PWADMM, sun2018markov}).
We call $(i_k)_{k\geq0}$ a walk of $\mathcal{G}$ if every $(i_k, i_{k+1}) \in \mathcal{E}$ 
and $z$ is passed as a token via the walk between agent $i_k$ and agent $i_{k+1}$.
Defining $\mathcal{N}_{i_k} (\mathcal{N}_{i_k} \subset \mathcal{N})$ the set of direct neighbors of agent ${i_k} (\in \mathcal{N})$ and $\overline{\mathcal{N}}_{{i_k}} = \mathcal{N}_{i_k} \cup \{{i_k}\}$, the probability of next activated agent ${i_{k+1}}$ can be denoted as $P_{i_{k},i_{k+1}} \in [0, 1]$, which is dependent on current agent $i_k$.
The implementation of proposed I-BCD is then presented in Algorithm \ref{alg1}.
Since only one agent and one link are active in each iteration, the convergence speed of IS is rather slow. 
To combat this problem, we propose to keep multiple agents and links active in an asynchronous mechanism.
\begin{algorithm}[t]
	\caption{Incremental Block-Coordinate Descent (I-BCD)} 
	\label{alg1}
	\begin{algorithmic}[1]
		\STATE \textbf{Initialize}: $\{x_i^0=\bm{0},z^0=\bm{0}|i\in\mathcal{N} \}$
		and agent $i_k$
		\FOR{$k =0,1,...$} 
		\STATE {receive} token $z^k$;
		\STATE {update} $x_{i}^{k+1}$ according to (\ref{eq:is_x});
		\STATE {update} $z^{k+1}$ according to (\ref{eq:is_z});
		\STATE {choose} $i_{k +1}(\in \overline{\mathcal{N}}_{i_{k}})$ according to $P_{i_{k},i_{k +1}}$;
		\STATE {send} $z^{k+1}$ to agent $i_{k+1}$.
    	\STATE \textbf{until} the stopping criterion is satisfied.
		\ENDFOR
	\end{algorithmic} 
\end{algorithm}

\section{Asynchronous Parallel Incremental Block-Coordinate Descent Framework} \label{sect:3}

To parallelize I-BCD and speed up the training process, we consider the asynchronous parallel incremental BCD (API-BCD) framework by allowing multiple walks to simultaneously train models among agents, in an asynchronous mode like \cite{liu2015asynchronous, peng2019convergence}.
To introduce API-BCD, we begin with a synchronous way with $M$ walks.
Denoting $z_m \in \mathbbm{R}^p$ as the $m$-th token $m \in \mathcal{M}= \{1,...,M\}$, problem (\ref{eq1}) is equal to
\begin{equation}\label{p:pro_mwalks} 
        \min_{\bm{x},\bm{z}} ~  \sum_{i=1}^{N}f_i(x_i); \quad s.t.~  z_m=x_i,~i\in\mathcal{N},~ m\in\mathcal{M} . 
\end{equation}
The corresponding penalty reformulation of (\ref{p:pro_mwalks}) can be represented as
\begin{equation}\label{p3}
    \min_{\bm x,\bm z}~ \mathcal{F}(\bm x, \bm{z}) := \sum_{i=1}^{N}f_i(x_i) + \frac{\tau}{2}\sum_{i=1}^{N}\sum_{m=1}^{M}\|x_i-z_m \|^2,
\end{equation}
where $\tau$ is the penalty parameter, $\bm{z} =  \{ z_1,...,z_M \} $.
Following the traditional synchronous mechanism \cite{dai2020synchronous}, updates of $\{ x_i \} \text{ and } \{ z_m \}$ at the $k$-th iteration follow
\begin{subequations}
    \begin{align}
        &x_i^{k+1} := \argmin_{x_i}f_i(x_i) + \frac{\tau}{2}\sum_{m=1}^{M}\|x_i - z_m^k \|^2,~i \in \mathcal{N};\label{eq:pis_x}\\
        &z_m^{k+1}:=\frac{1}{N}\sum_{i=1}^{N} x_i^{k+1},~m\in \mathcal{M}. \label{eq:pis_z}
    \end{align}
\end{subequations}
\begin{algorithm}[t]
	\caption{Asynchronous Parallel Incremental Block-Coordinate Descent (API-BCD) (logical view) } 
	\label{alg2}
	\begin{algorithmic}[1]
		\STATE \textbf{Initialize}: $\{x_i^0=\bm{0}, \hat{z}_{i,m}^0=\bm{0}|i\in\mathcal{N},m\in\mathcal{M} \}$ and $\{z_m^0=\bm{0},i_m \in\mathcal{M} \}$ \\
		\FOR{$k =0,1,...$} 
		\STATE receive token $z_m^k$ and set $\hat{z}_{i,m}\gets z_m$ with $i=i_{k}$ and $m = i_m$;
		\STATE {update} $x_i^{k+1}$ according to (\ref{eq:API-BCD_x});
		\STATE {update} $z_{i_m}^{k+1}$ according to (\ref{eq:API-BCD_z});
        \STATE {update} $\hat{z}_{i,m}^{k+1}$ according to (\ref{eq:API-BCD_local_z});
		\STATE {choose} $i_{k+1}(\in \overline{\mathcal{N}}_{i_{k}})$ according to $P_{{i_k}, {i_{k +1}}}$;
		\STATE {send} $z_{i_m}^{k+1}$ to agent $i_{k +1}$;
		\STATE \textbf{until} the stopping criterion is satisfied.
		\ENDFOR
	\end{algorithmic} 
\end{algorithm}
Note that agent $i$ needs to collect fresh $\{ z_m^{k} \}$ from all walks and then
$x_i^{k+1}$ is updated via (\ref{eq:pis_x}). 
Meanwhile, it is required to obtain fresh $\{ x_i^{k+1} \}$ from all the agents and then update $z_m^{k+1}$ by (\ref{eq:pis_z}).
{\color{black}{To make the update incremental and asynchronous, inspired by \cite{PWADMM}, we let each agent $i$ keep local copies $\{\hat{z}_{i, m} \}$ of $\{z_{m}\}$ for all $M$ walks.
Then, for active agent $i=i_k$,
we transform (\ref{eq:pis_x})-(\ref{eq:pis_z}) to the following processes for the updates of local model vector $x_i$, active token $z_{i_m}$ of the $i_m$-th walk and introduced local copies $\{\hat{z}_{i, m}\}$: }}
\begin{subequations}
    \begin{align}
        x_i^{k+1} :=& \argmin_{x_i}f_i(x_i) + \frac{\tau}{2}\sum_{m=1}^{M}\|x_i - \hat{z}_{i, m}^{k} \|^2,~i=i_k;\label{eq:API-BCD_x}\\
        z_{i_m}^{k + 1}:=& z_{i_m}^{k} + \frac{1}{N} ( x_{i_k}^{k + 1} - x_{i_k}^{k} ),~i_m\in \mathcal{M}; \label{eq:API-BCD_z}\\
         \hat{z}_{i, m}^{k+1} :=&
         \left \{   \begin{aligned}
         &z_{i_m}^{k + 1}, ~i=i_k, m=i_m; \\
         & z_{i_m}^{k},~\text{otherwise},\end{aligned} \right.\label{eq:API-BCD_local_z}
    \end{align}
\end{subequations}
where $z_{i_m}$\footnote{We note that $M$ walks are active simultaneously and each walk runs the procedure above on its own without any global synchronization. For convenience of the statement, we include a virtual counter $k$ to denote the iteration counter- every agent activates no matter on which agent will increase $k$ by $1$.} is updated in active agent ${i_k}$ and is then transferred to next active agent ${i_{k + 1}}$ via walk $i_m$.
\begin{figure} [t] 
	\vskip -0.1in
	\begin{center}
		\centerline{\includegraphics[width=82 mm]{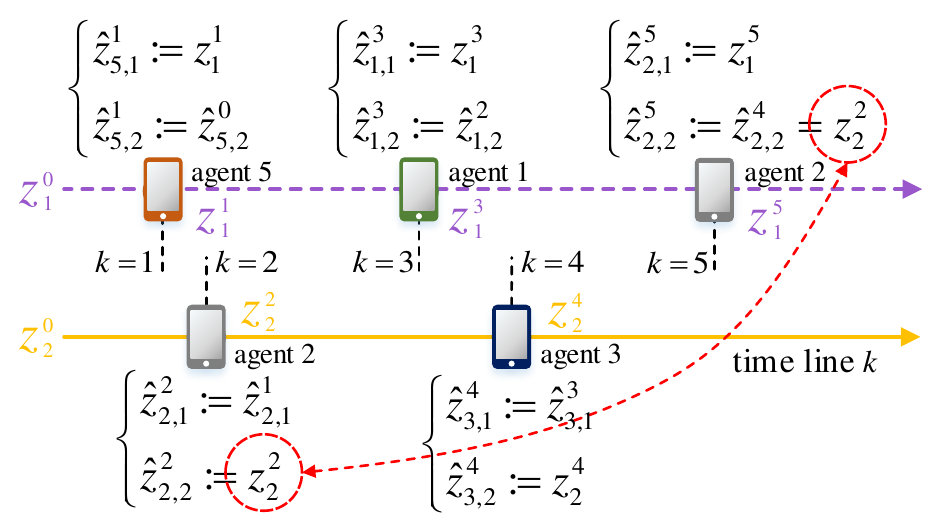}}
		\caption{Visualization of evolution of the local copy $\{z_m \}$ in a time line.}
	    \label{fig:update_order}
	\end{center}
	\vskip -0.4in
\end{figure}
We formally present proposed API-BCD in Algorithm \ref{alg2}.
Specifically, $M$ tokens are activated in parallel, each of which walks through the agents in the network with updating $x_i, z_m$ and $\hat{z}_{i, m}$ in agent $i_k$ and is further transferred to next active agent $i_{k +1}$ via steps 2-10 in Algorithm \ref{alg2}.
Similar to PW-ADMM in \cite{PWADMM}, transition of token $z_m$ follows the embedded Markov chain induced by the transmission matrix $\bm{P} \in \mathbbm{R}^{N\times N}$.
In Fig. \ref{fig:system}, we describe two walks of tokens for model training, where $z_1$ and $z_2$ model vectors move through a succession of agents in the network in parallel.
This reduces the idle time of each agent and the training time can be faster than that of the IS approach.
Fig. \ref{fig:update_order} also represents the evolution of the local copy $\{\hat{z}_{i, m}\}$ according to the described decentralized learning network in Fig. \ref{fig:system}.
In Fig. \ref{fig:update_order}, the horizontal line represents a time line.
Each agent is active at different stages of learning, such as agent 5 activated at time $k=1$ of the learning process, where local copy $\hat{z}_{5, 1}$ is updated by token $z_1$.
Similarly, agent 1 updates local copy $\hat{z}_{1,1}$ and agent 3 updates local copy $\hat{z}_{3,2}$ by token $z_1$ and $z_2$, respectively.
Specially, in agent 2, both local copies $\{ \hat{z}_{2,m} \}_{m=1}^2$ are updated although these two updates happen in two different time (i.e., time $k=2$ and time $k=5$).

}}



\section{Convergence Analyses}\label{sec:convergence}
In this section, we analyze the convergence properties of the proposed I-BCD and API-BCD algorithms over a decentralized ML system.
Specifically, we establish the convergence of both IS and API-BCD approaches in convex scenarios.
Suppose that local loss $f_i(x_i)$ of each agent is convex and bounded from below.
Then, we have the following results.

\begin{theorem}\label{theorem_1}
	Let $(\bm x^k,z^k)$ be the iterates generated by I-BCD approach.
	For active agent $i_k$ at iteration round $k$, it yields the following descent
	\begin{equation}
	    \begin{aligned}
	    &\mathcal{F}(\bm x^{k+1}, z^{k+1})  - \mathcal{F}(\bm x^k, z^k)  \\
	    &~~~~~~~~~~\leqslant  -  \frac{\tau}{2}  \| x_{i_k}^{k+1} - x_{i_k}^k \|^2 - \frac{\tau N}{2} \| z^{k+1} - z^k\|^2.
	\end{aligned}
	\end{equation} 
\end{theorem}
\begin{IEEEproof}
The proof of Theorem \ref{theorem_1} can be obtained in Appendix \ref{firstAppendix}.
\end{IEEEproof}
Though API-BCD is an asynchronous approach, we prove the convergence from the perspective of synchronous point of view by allowing fresh tokens sharing, that is, all agents share fresh $\{ z_m  \}$.
Then, we obtain the following result.
\begin{theorem}\label{theorem_2}
Let $(\bm x^k, \bm z^k)$ be the iterates generated by {API-BCD} approach with sharing fresh tokens $\{ z_{m}  \}$.
For active agent $i_k$ at the iteration round $k$, it yields the following descent
\begin{equation}
    \begin{aligned}
		    & \mathcal{F}(\bm x^{k+1},\bm{z}^{k+1}) - \mathcal{F}(\bm x^k,\bm{z}^k)  \\
		    &~~~~~\leqslant  - \frac{\tau M }{2} \|x_{i_k}^{k+1} - x_{i_k}^k \|^2 - \frac{\tau N}{2} \sum_{m=1}^M \| z_m^{k+1} - z_m^k\|^2.
		\end{aligned}
\end{equation}
		
\end{theorem}
\begin{IEEEproof}
The proof of Theorem \ref{theorem_2} can be obtained in Appendix \ref{secondAppendix}.
\end{IEEEproof}

\begin{remark}
It is mentioned that the API-BCD algorithm can be extended to a gradient-based variant to reduce computational complexity by adopting the first-order approximation as well as a quadratic proximal term to stabilize the convergence behavior for update in (\ref{eq:API-BCD_x}).
\end{remark}
In what follows, we will analyze the convergence of gradient-based API-BCD (gAPI-BCD) with first-order approximation in $f_i(x_i)$, along with a quadratic proximal term with parameter $\rho$ in $x_i$.
Denoting by $\nabla f_i(x_i^k)$ the local gradient in agent $i$, the update of local model vector $x_i$ at the $k$-th iteration in gAPI-BCD follows
\begin{equation}
  \begin{aligned}
     x_i^{k+1} :=\argmin_{x_i}~  & \langle \nabla f_i(x_i^k), x_i - x_i^k \rangle  + \frac{\tau}{2}\sum_{m=1}^{M}\|x_i - \hat{z}_{i, m}^{k} \|^2 \\
    & + \frac{\rho}{2} \|x_i - x_i^k\| ^2,~i=i_k,\label{eq:gAPI-BCD}
\end{aligned}  
\end{equation} 
while updates of active token $z_{i_m}$ of the $i_m$-th walk and local copies $\{\hat{z}_{i, m}\}$ keep the same compared with API-BCD.
We first introduce Assumption \ref{ass:smooth}, which is widely adopted in the analysis of decentralized ML and convex optimization.
\begin{assumption}[Smoothness]
    \label{ass:smooth}
	The local loss function $f_i({x})$ is \textit{L-}smooth, i.e., for any ${x}, {y} \in\mathbbm{R}^{p}$,
\begin{equation}  \label{ass:smooth_eq1}
\begin{aligned}
\norm{ \nabla f_i({x}) -  \nabla f_i({y}) } \leqslant L \norm{  {x} -{y}  }, \forall i \in \mathcal{N},
\end{aligned}
\end{equation}
and this is equivalent to 
\begin{equation} \label{ass:smooth_eq2}
f_i({x}) \leqslant f_i({y}) + \left\langle  \nabla f_i({y}), {x}-{y}  \right\rangle + \frac{L}{2} \norm{{x}-{y}}^2, \forall i \in \mathcal{N}.
\end{equation}
\end{assumption}

\begin{theorem} \label{theorem_3}
Suppose Assumption \ref{ass:smooth} holds and
let $(\bm x^k, \bm z^k)$ be the iterates generated by {gAPI-BCD} approach with sharing fresh tokens $\{ z_{m}  \}$, active agent $i_k$ at the iteration round $k$ yields the following descent
\begin{equation}
    	\begin{aligned}
		      &\mathcal{F}(\bm x^{k+1},\bm{z}^{k+1}) - \mathcal{F}(\bm x^k,\bm{z}^k)  \\
		    &~~~~~~~~~~~ \leqslant- ( \frac{\tau M }{2} +\rho -\frac{L}{2} ) \|x_{i_k}^{k+1} - x_{i_k}^k \|^2  \\
		    &  ~~~~~~~~~~~~~~~  - \frac{\tau N}{2} \sum_{m=1}^M \| z_m^{k+1} - z_m^k\|^2.
		\end{aligned}
\end{equation}
	
\end{theorem}
\begin{IEEEproof}
The proof of Theorem \ref{theorem_3} is similar to that of Theorem 2. By substituting {\color{black} (\ref{eq:term_e}) with (\ref{ass:smooth_eq2})} and the optimality condition $\nabla f_{i_k}(x_{i_k}^{k+1}) + \tau \sum_{m=1}^M  ( x_{i_k}^{k+1} - z_m^k ) +\rho  ( x_{i_k}^{k+1} - x_{i_k}^k  ) = 0$, Theorem \ref{theorem_3} can be obtained.
\end{IEEEproof}
\begin{remark}
Theorem \ref{theorem_1}, Theorem \ref{theorem_2} and Theorem \ref{theorem_3} state that the objective function values converge.
We note that the convergence performance of API-BCD using local copies $\{\hat{z}_{i,m}\}$ will be much more complicated.
\end{remark}

\section{Numerical results} \label{sec:4}

 \begin{figure}[] 
     \vskip -0.1 in
     	\centering
     	\subfloat[]{\hspace*{1 mm}\includegraphics[width=60 mm]{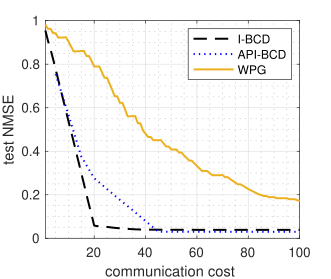}\label{fig_c}}\\
     	\vskip -0.02 in
     	\subfloat[]{\hspace*{1 mm}\includegraphics[width=60 mm]{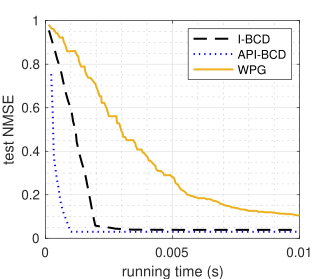}\label{fig_d}} 
 	\vskip  -0.1 in 
     	\caption{Test NMSE on \textit{cpusmall} dataset with $N=20,\zeta=0.7,K=5,\alpha=0.5, \tau _{\text{IS}}=1,\tau _{\text{API-BCD}}=0.1$. }\label{fig:fig_cpusmall} 
\end{figure}

 \begin{figure}[] 
     \vskip -0.1 in
     	\centering
     	\subfloat[]{\hspace*{1mm}\includegraphics[width =60 mm]{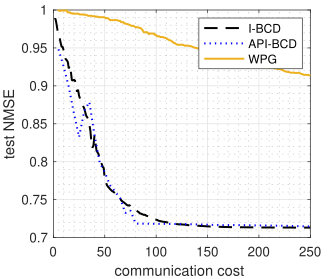}}
     	\\
     	\vskip -0.02 in
     	\subfloat[]{\hspace*{1mm}\includegraphics[width=60 mm]{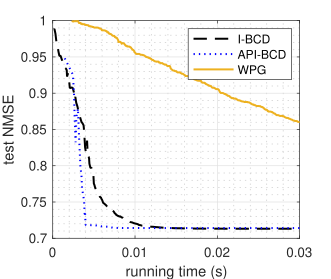}\label{fig_b}} 
 	    \vskip  -0.1 in 
     	\caption{Test NMSE on \textit{cadata} dataset with $N=50,\zeta=0.7,K=5,\alpha=0.2,\tau _{\text{IS}}=2.8,\tau _{\text{API-BCD}}=0.1$.}\label{fig:fig_cadata} 
\end{figure}

 \begin{figure}[] 
     \vskip -0.1 in
     	\centering
     	\subfloat[]{\hspace*{1 mm}\includegraphics[width =60 mm]{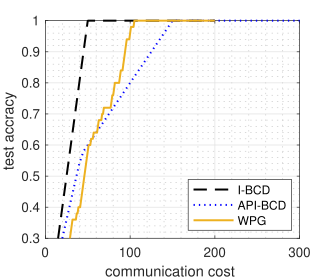}}\\
     	\vskip -0.02 in
     	\subfloat[]{\hspace*{1 mm}\includegraphics[width=60 mm]{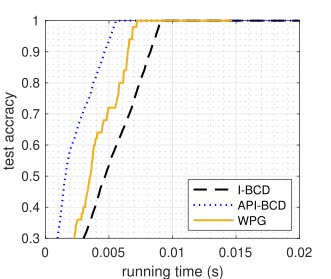}\label{fig_b}} 
 	\vskip  -0.1 in 
     	\caption{Test accuracy on \textit{ijcnn1} dataset with $N=50,\zeta=0.7,K=5,\alpha=0.5,\tau _{\text{IS}}=2.8,\tau _{\text{API-BCD}}=0.1$. }\label{fig:fig_ijcnn1}
\end{figure}

 \begin{figure}[] 
     \vskip -0.1 in
     	\centering
     	\subfloat[]{\hspace*{1 mm}\includegraphics[width=60 mm]{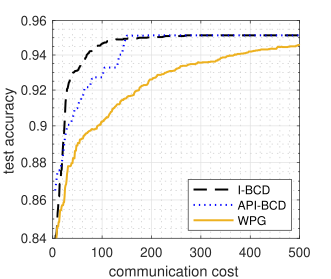}\label{fig_c}}\\
     	\vskip -0.02 in
     	\subfloat[]{\hspace*{1 mm}\includegraphics[width=60 mm]{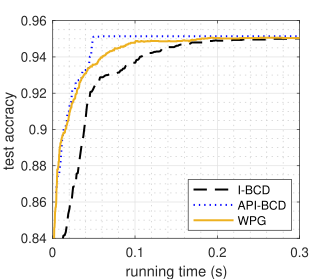}\label{fig_d}} 
 	\vskip  -0.1 in 
     	\caption{Test accuracy on \textit{USPS} dataset with $N=10,\zeta=0.7,K=5,\alpha=0.1,\tau _{\text{IS}}=5,\tau _{\text{API-BCD}}=1$. }\label{fig:fig_mnist}
\end{figure}

{\color{black}
We perform experimental simulations to verify the convergence performance of the proposed approaches: I-BCD in Algorithm \ref{alg1} and API-BCD in Algorithm \ref{alg2} and evaluate the learning performance for decentralized systems under four real datasets, \textit{cpusmall}  \cite{dataset_libsvm}, \textit{cadata}  \cite{dataset_libsvm}, \textit{ijcnn1} \cite{dataset_libsvm}, and \textit{USPS} \cite{dataset_mnist}. 
For simulations, the considered connected network $\mathcal{G}$ consists of $N$ agents and $|\mathcal{E}|=\frac{N(N-1)}{2}\xi$ links. 
We consider unicast among agents, and $M$ walks are activated for API-BCD. 
The communication cost per link is $1$ unit and 
the consumed time for each communication is assumed to follow $\mathcal{U}(10^{-5},10^{-4})$ s.
The running time is measured by both computation time in local agents and communication time between agents.
To investigate the communication and running time costs, we compare our approaches with one typical incremental method WPG \cite{WPG}, where agents are activated in a predefined circulant pattern, i.e., Hamiltonian cycle and the model is periodically trained according to
\begin{equation}
 \begin{aligned}
    &x_i^{k+1} := z^k - \alpha \nabla f_i(z^k),~i= i_k; \\
    &z^{k+1} := z^k + \frac{1}{N}  ( x_{i_k}^{k+1} - x_{i_k}^k ),
\end{aligned}   
\end{equation} 
where $z^{k+1}$ is updated in active agent $i_k$ and passed as a token to the next agent, $\alpha$ is the learning rate and $ \nabla f_i(z^k)$ is the gradient with respect to its local dataset.
For a fair comparison, in the following we shall concentrate on a deterministic agent selection rule similar to \cite{WPG}.

We focus on two kinds of decentralized ML tasks, i.e., training a linear regression model and a classification model under different network typologies.
The first set of experiments considers training a least square problem.
In Fig. \ref{fig:fig_cpusmall}, the normalized mean square error (NMSE) performance of the proposed approaches is presented in terms of both running time and communication cost.
It is obvious that the proposed API-BCD on \textit{cpusmall} dataset is the most efficient in running time.
This is because that the inherent asynchronous mechanism of API-BCD outperforms the synchronous-based approaches.
Meanwhile, the result in Fig.3(a) shows that our proposed incremental-based learning approaches (i.e., I-BCD and API-BCD) consume less communication costs compared to that of WPG approach.
In a larger test network, Fig. \ref{fig:fig_cadata} presents experimental results based on \textit{cadata} dataset and the similar learning performance is observed as well.
In Fig. \ref{fig:fig_ijcnn1} and Fig. \ref{fig:fig_mnist}, we also evaluate the performance of proposed approaches in classification problems over both communication and time costs.
As shown in Fig. \ref{fig:fig_ijcnn1} on \textit{ijcnn1} dataset, compared to both I-BCD and WPG approaches, the proposed API-BCD algorithm can guarantee both communication-efficiency and fast convergence.
In Fig. \ref{fig:fig_mnist}, the curves under different network setups present the similar trends as those of Fig. \ref{fig:fig_ijcnn1}.

}

\section{Conclusions}
\label{sect:5}
We have developed incremental-based frameworks for decentralized learning to accelerate learning process in terms of both communication and running time costs.
The convergence of our proposed approaches are theoretically analyzed.
By numerical experiment, we show that the proposed API-BCD is effective in achieving target accuracy with less time consumption as well as comparable communication cost.
Moreover, the proposed approach is flexible and scalable in terms of network size.

\appendices
\section{Proof to Theorem \ref{theorem_1}} \label{firstAppendix}
\begin{IEEEproof}
We start from convexity property. 
Using convexity of $f_i(x_i)$, we can drive
\begin{align} \label{eq:convexity}
    f_i(x_i^{k+1}) - f_i (x_i^k) \leq  \langle \nabla f_i(x_i^{k+1}), x_i ^{k+1} - x_i^k  \rangle,
\end{align}
where $\nabla f_i(x_i^{k+1})$ is the gradient.
Remember agent $i_k$ is activated at iteration $k$. Then, the optimality condition of (\ref{eq:is_x}) for $i=i_k$ implies
\begin{align} \label{eq:optimal_condition_x}
    \nabla f_{i_k}(x_{i_k}^{k+1}) + \tau ( x_{i_k}^{k+1} - z^k ) = 0.
\end{align}
After updating to $\bm{x}^{k+1}$ and $z^{k+1}$ by (\ref{eq:is_x}) and (\ref{eq:is_z}), we have 
\begin{equation}
 \begin{aligned} \label{eq:loss_des1}
    &\mathcal{F}( \bm x^{k+1}, z^{k+1}) - \mathcal{F}( \bm x^k, z^k) = \underline{f_{i_k}(x_{i_k}^{k+1}) - f_{i_k} (x_{i_k}^k)}_{A}  \\
    &  \qquad + \underline{\frac{\tau}{2} \sum _{i=1} ^N \| x_i^{k+1} - z^{k+1} \|^2 - \frac{\tau}{2} \sum _{i=1} ^N \|x_i^{k} - z^{k} \|^2}_{B}.
\end{aligned}   
\end{equation} 
Term $A$ in (\ref{eq:loss_des1}) is upper bounded by
\begin{align}
     A &\leq  \langle \nabla f_{i_k}(x_{i_k}^{k+1}), x_{i_k} ^{k+1} - x_{i_k}^k  \rangle  
       =  \tau  \langle x_{i_k}^{k+1} - z^k, x_{i_k}^{k} - x_{i_k}^{k+1}  \rangle,
\end{align}
due to (\ref{eq:optimal_condition_x}) and (\ref{eq:convexity}),
whilst term $B$ satisfies
\begin{align} \label{eq:term_b}
    B &= \frac{\tau}{2} \sum _{i=1} ^N  ( \| x_i^{k+1} - z^{k} + z^{k} - z^{k+1} \|^2 - \| x_i^{k} - z^{k} \|^2  )\notag\\
    &= \frac{\tau}{2} \sum _{i=1}^N   ( \| x_i^{k+1}- z^{k} \|^2 - \| x_i^{k} - z^{k} \|^2 + \| z^{k+1} - z^k \|^2   \notag\\
    &\qquad\qquad~ +   2 \langle x_i^{k+1} - z^k, z^k - z^{k+1}  \rangle )    \notag\\
    &= \underline{\frac{\tau}{2}\sum _{i=1}^N    ( \| x_i^{k+1}- z^{k} \|^2 - \| x_i^{k} - z^{k} \|^2  ) }_C\notag\\
    &\quad + \frac{\tau}{2}\sum _{i=1}^N \| z^{k+1} - z^k \|^2 + \underline{\tau \sum _{i=1}^N \langle x_i^{k+1} - z^k, z^k - z^{k+1} \rangle }_D.
\end{align}
Term $C$ in (\ref{eq:term_b}) can be expressed as
\begin{equation}
  \begin{aligned}
   C& \overset{(a)}{=} \frac{\tau}{2} \sum _{i=1}^N  ( \| x_{i}^{k+1}- x_{i}^{k} \|^2 + \tau  \langle   x_i^{k} - z^k, x_i^{k+1} - x_i^{k} \rangle   ) \\
    &=\frac{\tau}{2} \|x_{i_k}^{k+1}- x_{i_k}^{k} \|^2 + \tau \langle   x_{i_k}^{k} - z^k, x_{i_k}^{k+1} - x_{i_k}^{k}  \rangle,
\end{aligned}  
\end{equation} 
where $(a)$ holds since the cosine identity $\norm{b+c}^2- \norm{a+c}^2 = \norm{b-a}^2+ 2\left\langle a+c, b-a \right\rangle$.
Term $D$ can be also transferred as
\begin{equation}
   \begin{aligned} \label{eq:term_d}
    D &=\tau  \langle \sum _{i=1}^N  ( x_i^{k+1} - z^k  ), z^k - z^{k+1} \rangle  \\
    &=\tau N  \langle \frac{1}{N}\sum _{i=1}^N x_i^{k+1} - z^{k}, z^k - z^{k+1}  \rangle  \\
    &\overset{(b)}{=}\tau N  \langle z^{k+1} - z^k, z^k - z^{k+1}  \rangle \\
    &= -\tau N \| z^{k+1} - z^k \|^2,
\end{aligned} 
\end{equation} 
where (b) is from $z^{k+1} = \frac{1}{N} \sum _{i=1}^N x_i^{k+1}$.
Thus, (\ref{eq:loss_des1}) can be derived as
\begin{equation}
  \begin{aligned}
    &\mathcal{F}(\bm x^{k+1},z^{k+1}) - \mathcal{F}(\bm x^k, z^k) \\
    &\leq - \tau \langle x_{i_k}^{k+1} - z^k, x_{i_k}^{k+1} - x_{i_k}^{k} \rangle + \frac{\tau}{2} \| x_i^{k+1}- x_i^{k} \|^2 \\
    &\quad + \tau \langle   x_{i_k}^{k} - z^k, x_{i_k}^{k+1} - x_i^{k}\rangle + \frac{\tau}{2}\sum _{i=1}^N \| z^{k+1} - z^k \|^2  \\
    & \quad -\tau N \| z^{k+1} - z^k \|^2 \\
    &= - \frac{\tau}{2} \| x_{i_k}^{k+1} - x_{i_k}^{k}\|^2 - \frac{\tau N}{2} \| z^{k+1} - z^k \|^2.
\end{aligned}  
\end{equation} 
This completes the proof of Theorem 1.
\end{IEEEproof}

\section{Proof to Theorem \ref{theorem_2}} \label{secondAppendix}
\begin{IEEEproof}
Here, we consider the convergence of the API-BCD method with sharing fresh tokens $\{ z_m\}$.
Similar to (\ref{eq:optimal_condition_x}), the optimality condition of (\ref{eq:API-BCD_x}) for $i=i_k$ implies
\begin{align} \label{eq:optimal_condition_api_bcd}
    \nabla f_{i_k}(x_{i_k}^{k+1}) + \tau \sum_{m=1}^M  ( x_{i_k}^{k+1} - z_m^k ) = 0,
\end{align}
since $\hat{z}_{i, m}^t = z_m^k, \forall i \in  \mathcal{N} $ is satisfied.
Then, the descent of objective function follows
\begin{align} \label{eq:loss_descent_2}
    &\mathcal{F}(\bm x^{k+1}, \bm z^{k+1}) - \mathcal{F}(\bm x^k, \bm z^k)  
    =  \underline{f_{i_k}(x_{i_k}^{k+1}) - f_{i_k} (x_{i_k}^k)}_E \notag\\
    &~~ + \underline{\frac{\tau}{2} \sum _{i=1} ^N \sum_{m=1}^M \| x_i^{k+1} - z_m^{k+1} \|^2 - \frac{\tau}{2} \sum _{i=1} ^N \sum_{m=1}^M  \| x_i^{k} - z_m^{k} \|^2}_F.
\end{align}
Following (\ref{eq:convexity})-(\ref{eq:term_d}), term ${E}$ and term ${F}$ in (\ref{eq:loss_descent_2}) can be derived as
\begin{equation}
    \begin{aligned} \label{eq:term_e}
    E &\leq  \langle \nabla f_{i_k}(x_{i_k}^{k+1}), x_{i_k} ^{k+1} - x_{i_k}^k  \rangle  \\
    &\overset{(c)}{=}- \tau  \langle \sum _{m=1}^M  ( x_{i_k}^{k+1} - z_m^k  ), x_{i_k}^{k+1} - x_{i_k}^{k}  \rangle  \\
    &= - \tau \sum _{m=1}^M  \langle x_{i_k}^{k+1} - z_m^k, x_{i_k} ^{k+1} - x_{i_k}^k  \rangle  \\
    &= - \tau \sum _{i=1}^N  \sum _{m=1}^M  \langle x_{i}^{k+1} - z_m^k, x_{i} ^{k+1} - x_{i}^k  \rangle,
\end{aligned}
\end{equation} 
and
\begin{equation}
    \begin{aligned}
    F &= \frac{\tau}{2} \sum _{i=1}^N  \sum _{m=1}^M   ( \| x_i^{k+1} - z_m^{k} + z_m^{k} - z_m^{k+1} \|^2 - \| x_i^{k} - z_m^{k} \|^2  ) \\
    &= \frac{\tau}{2} \sum _{i=1}^N  \sum _{m=1}^M   ( \| x_i^{k+1} - z_m^{k} \|^2 -  \| x_i^{k} - z_m^{k} \|^2 + \| z_m^{t+1} - z_m^{t} \|^2  \\
    & \qquad\qquad\qquad ~   + 2 \langle x_i^{k+1} - z_m^{k},  z_m^{k} - z_m^{k+1}   \rangle  )  \\
    &\overset{(d)}{=}\frac{\tau}{2} \sum _{i=1}^N  \sum _{m=1}^M  \| x_i^{k+1} -  x_i^{k} \|^2 + \frac{\tau}{2} \sum _{i=1}^N  \sum _{m=1}^M  \| z_m^{k} - z_m^{k+1} \|^2  \\
    & ~\quad + \underline{\tau \sum _{i=1}^N  \sum _{m=1}^M \langle x_i^{k+1} - z_m^k, z_m^k - z_m^{k+1}  \rangle}_G  \\
    & ~\quad + \tau \sum _{i=1}^N  \sum _{m=1}^M  \langle x_i^{k} - z_m^{k},  x_i^{k+1} -  x_i^{k} \rangle,
\end{aligned}
\end{equation} 
where $(c)$ is due to (\ref{eq:optimal_condition_api_bcd}) and $(d)$ is from cosine identity, respectively.
Meanwhile, term $G$ can be derived as
\begin{equation}
  \begin{aligned}
    G
    &= \tau N \sum _{m=1}^M  \langle \frac{1}{N} \sum _{i=1}^N x_i^{k+1} - z_m^k, z_m^k - z_m^{k+1}  \rangle  \\
    &= \tau N \sum _{m=1}^M   \langle \frac{1}{N} \sum _{i=1}^N x_i^{k+1} - z_m^{k+1} + z_m^{k+1} - z_m^k, z_m^k - z_m^{k+1}  \rangle  \\
    &= \tau N \sum _{m=1}^M  \langle \frac{1}{N} \sum _{i=1}^N x_i^{k+1} - z_m^{k+1}, z_m^k - z_m^{k+1}  \rangle  \\
    &\quad - \tau N \sum _{m=1}^M \| z_m^{k+1} - z_m^k \| ^2  \\
    &\overset{(e)}{=} - \tau N \sum _{m=1}^M \| z_m^{k+1} - z_m^k \| ^2,
\end{aligned}   
\end{equation} 
where (e) holds because of (\ref{eq:pis_z}). 
Hence, (\ref{eq:loss_descent_2}) is equivalently equal to 
\begin{equation}
    \begin{aligned}
    &\mathcal{F}(\bm x^{k+1},\bm z^{k+1}) - \mathcal{F}(\bm x^k, \bm z^k)  \\
    \leq& - \tau \sum _{i=1}^N  \sum _{m=1}^M  \langle x_{i}^{k+1} - z_m^k, x_{i} ^{k+1}     - x_{i}^k  \rangle  \\
    & +\frac{\tau M}{2} \sum _{i=1}^N \| x_i^{k+1} -  x_i^{k} \|^2 - \tau N \sum _{m=1}^M \| z_m^{k+1} - z_m^k \| ^2 \\
    & + \frac{\tau N}{2} \sum _{m=1}^M  \| z_m^{k} - z_m^{k+1} \|^2 + \tau \sum _{i=1}^N  \sum _{m=1}^M  \langle x_i^{k} - z_m^{k},  x_i^{k+1} -  x_i^{k} \rangle \\
    =&   ( -\tau M + \frac{\tau M}{2}  ) \sum _{i=1}^N \|x_i^{k+1} -  x_i^{k} \|^2 - \frac{\tau N}{2} \sum _{m=1}^M  \| z_m^{k+1} - z_m^k \| ^2 \\
    =&- \frac{\tau M}{2} \|x_{i_k}^{k+1} -  x_{i_k}^{k} \|^2 - \frac{\tau N}{2} \sum _{m=1}^M \| z_m^{k+1} - z_m^k \| ^2,
\end{aligned}
\end{equation} 
which completes the proof of Theorem 2.
\end{IEEEproof}


\ifCLASSOPTIONcompsoc
  \section*{Acknowledgments}
\else
  \section*{Acknowledgment}
\fi

This work was supported by ERA-NET Smart Energy Systems SG+ 2017 Program, "SMART-MLA" with Project number 89029 (and SWEA number 42811-2), Swedish Research Council Project entitled "Coding for Large-scale Distributed Machine Learning", Swedish Foundation for International Cooperation in Research and Higher Education (STINT), project "Efficient and Secure Distributed Machine Learning with Gradient Descend", and FORMAS project entitled "Intelligent Energy Management in Smart Community with Distributed Machine Learning", number 2021-00306. 

\bibliography{reflibb}
\bibliographystyle{IEEEtran} 

\end{document}